\documentclass[runningheads]{llncs}

\usepackage{amsmath,amssymb,amsfonts}
\usepackage{graphicx}
\usepackage{booktabs}
\usepackage{multirow}
\usepackage{bm}
\usepackage{float}
\usepackage{placeins}
\usepackage[pagebackref,breaklinks,colorlinks,citecolor=blue]{hyperref}
\usepackage{cleveref}

\begin{document}
\emergencystretch=1.5em

\title{Deformation-Free Cross-Domain Image Registration via Position-Encoded Temporal Attention}
\titlerunning{Deformation-Free Cross-Domain Registration via Position-Encoded Temporal Attention}

\author{Yiwen Wang \and Jiahao Qin\thanks{Corresponding author: jiahao.qin19@gmail.com}}
\authorrunning{Y. Wang, J. Qin}
\institute{}

\maketitle

%==============================================================================
\begin{abstract}
\emergencystretch=1em
We address the problem of cross-domain image registration, where paired images $I_m$ and $I_f$ exhibit coupled geometric misalignment and domain-specific appearance shift. We formalize this as a factorization problem: decomposing each image into a domain-invariant scene representation $\bm{s} \in \mathbb{R}^{C \times H \times W}$ and a global appearance statistic $\bm{a} \in \mathbb{R}^{d}$, such that registration reduces to recombining the scene structure of $I_m$ with the appearance of $I_f$ via Adaptive Instance Normalization (AdaIN). This factorization eliminates the need for explicit deformation field estimation. To exploit temporal coherence in sequential acquisitions, we introduce a position-encoded cross-frame attention mechanism that fuses learnable and sinusoidal position embeddings with multi-head attention over a sliding window of $k$ neighboring frames, enriching the scene representation with inter-frame context. We instantiate this framework as GPEReg-Net and evaluate on two benchmarks: FIRE-Reg-256 (retinal fundus, semi-rigid) and HPatches-Reg-256 (synthetic textured patches, affine). GPEReg-Net achieves state-of-the-art performance on both benchmarks (FIRE: SSIM\,=\,0.928, PSNR\,=\,33.47\,dB; HPatches: SSIM\,=\,0.450, PSNR\,=\,21.01\,dB), surpassing all baselines including deformation-based methods, while running $1.87\times$ faster than SAS-Net.
Code: \url{https://github.com/JiahaoQin/GPEReg-Net}.

\end{abstract}

%==============================================================================
\section{Introduction}
\label{sec:introduction}
%==============================================================================

Let $I_m, I_f \in \mathbb{R}^{H \times W}$ denote a moving image and a fixed image, respectively. Image registration seeks a mapping $\mathcal{T}: I_m \mapsto \hat{I}_r$ such that $\hat{I}_r \approx I_f$ under a suitable similarity measure~\cite{sotiras2013deformable,chen2024survey}. In cross-domain settings---where $I_m$ and $I_f$ are drawn from different intensity distributions due to varying acquisition conditions (e.g., subject motion in retinal imaging~\cite{hernandez2017fire}, viewpoint changes in natural images~\cite{balntas2017hpatches})---the brightness constancy assumption $I_m(\bm{x}) \approx I_f(\bm{x} + \bm{u})$ underlying conventional methods~\cite{chen2024survey} is systematically violated.

\textbf{Limitations of existing approaches.}
Classical methods---SIFT~\cite{lowe2004distinctive}, Demons~\cite{vercauteren2009diffeomorphic}, optical flow~\cite{horn1981determining}, SyN~\cite{avants2008symmetric}---estimate spatial correspondences directly but degrade when intensity distributions differ. Learned deformation methods (VoxelMorph~\cite{balakrishnan2019voxelmorph}, TransMorph~\cite{chen2021transmorph}, spatial transformer networks~\cite{jaderberg2015spatial}, Laplacian pyramid approaches~\cite{mok2020large}) parameterize $\mathcal{T}$ as a displacement field $\bm{u}(\bm{x})$ but inherit the same intensity assumption. Scene-appearance separation frameworks~\cite{qin2026sasnet,qin2026domainreg} address domain shift through disentangled latent spaces but employ complex generative architectures without temporal awareness. Progressive strategies~\cite{qin2026progressive} improve alignment iteratively but do not explicitly factorize appearance from scene content.

\textbf{Key insight.}
We observe that the cross-domain registration problem admits a natural \emph{factorization}: each image can be decomposed into a domain-invariant scene representation $\bm{s}$ (encoding spatial structure) and a domain-specific appearance statistic $\bm{a}$ (encoding intensity profile). Registration then reduces to recombining $\bm{s}_m$ with $\bm{a}_f$ via Adaptive Instance Normalization (AdaIN)~\cite{huang2017adain}, yielding $\hat{I}_r = \mathcal{D}(\text{AdaIN}(\bm{s}_m, \bm{a}_f))$ without estimating any deformation field~\cite{qin2023atd,qin2024zoom,qin2025bcpmjrs}. Furthermore, in sequential acquisitions, temporal coherence between consecutive frames can be exploited via position-encoded cross-frame attention~\cite{vaswani2017attention,qin2025ancestral,qin2024biomamba} to improve inter-frame consistency.

We instantiate this insight as \textbf{GPEReg-Net} and make the following contributions:

\begin{enumerate}
    \item \textbf{Scene-appearance factorization.} We formalize cross-domain registration as a latent factorization problem and propose an encoder-decoder architecture that decomposes images into domain-invariant scene features $\bm{s} \in \mathbb{R}^{64 \times H \times W}$ and global appearance codes $\bm{a} \in \mathbb{R}^{32}$, with AdaIN-based recombination that eliminates deformation field estimation entirely.

    \item \textbf{Position-encoded temporal attention.} We introduce a Global Position Encoding (GPE) module that fuses learnable position embeddings, sinusoidal encoding, and multi-head cross-frame attention over a sliding window of $k$ neighbors, enabling the model to exploit temporal structure in sequential acquisitions.

    \item \textbf{Comprehensive cross-domain evaluation.} We validate on two diverse benchmarks---FIRE-Reg-256 (retinal fundus, semi-rigid) and HPatches-Reg-256 (synthetic patches, affine)---achieving state-of-the-art results on both while maintaining real-time throughput ($69$ FPS).
\end{enumerate}

%==============================================================================
\section{Proposed Method: GPEReg-Net}
\label{sec:method}
%==============================================================================

\subsection{Problem Formulation and Architecture Overview}
\label{sec:overview}

Given a moving image $I_m \in \mathbb{R}^{H \times W}$ and a fixed image $I_f \in \mathbb{R}^{H \times W}$ drawn from potentially different intensity distributions, our goal is to learn a mapping $\mathcal{T}_\theta: (I_m, I_f) \mapsto \hat{I}_r$ such that $\hat{I}_r$ is geometrically aligned with $I_f$ while adopting its intensity profile, \emph{without} estimating an explicit deformation field $\bm{u}: \mathbb{R}^2 \to \mathbb{R}^2$.

\begin{figure}[!htb]
\centering
\includegraphics[width=\linewidth]{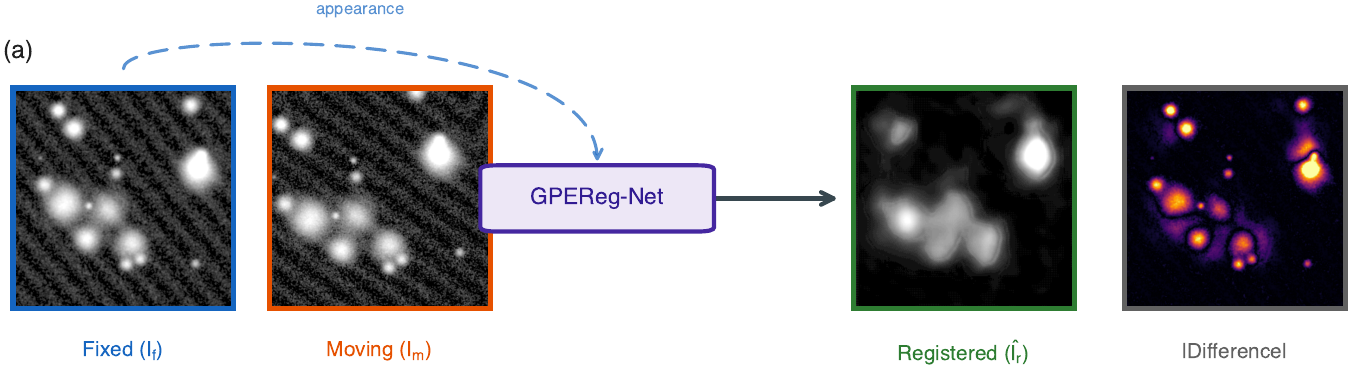}\\[4pt]
\includegraphics[width=\linewidth]{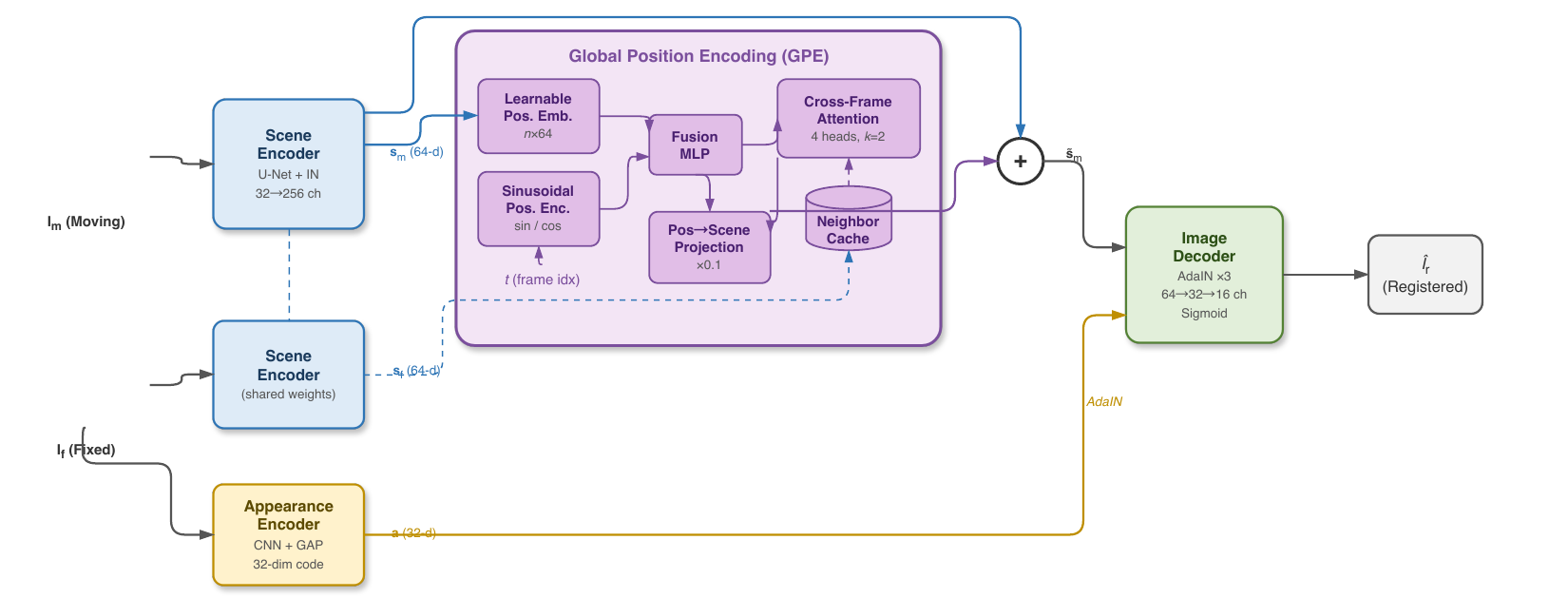}
\caption{(a) Registration example on HPatches-Reg-256~\cite{balntas2017hpatches}: the fixed image $I_f$, moving image $I_m$ with geometric misalignment, the registered output $\hat{I}_r$ produced by GPEReg-Net, and the absolute difference $|\hat{I}_r - I_f|$. A dashed arrow indicates the appearance pathway from $I_f$. (b) Overview of GPEReg-Net. The SceneEncoder $\mathcal{S}$ extracts domain-invariant features $\bm{s}$ from the moving image $I_m$ using instance normalization. The AppearanceEncoder $\mathcal{A}$ extracts a 32-dim global appearance code $\bm{a}$ from the fixed image $I_f$. The GPE module $\mathcal{G}$ enriches scene features with temporal context via learnable position embeddings, sinusoidal encoding, and cross-frame attention, producing $\tilde{\bm{s}}$. The ImageDecoder $\mathcal{D}$ uses AdaIN to modulate enhanced scene features with the target appearance, producing the registered output $\hat{I}_r$ without any explicit deformation field.}
\label{fig:architecture}
\end{figure}

We decompose $\mathcal{T}_\theta$ into four learned modules (\cref{fig:architecture}):

\begin{enumerate}
    \item \textbf{SceneEncoder} $\mathcal{S}$: Extracts domain-invariant scene features $\bm{s} = \mathcal{S}(I_m)$ via instance normalization, discarding appearance information.

    \item \textbf{AppearanceEncoder} $\mathcal{A}$: Extracts a global appearance code $\bm{a} = \mathcal{A}(I_f)$ that captures the intensity profile of the target domain.

    \item \textbf{Global Position Encoding} $\mathcal{G}$: Enhances scene features with temporal context: $\tilde{\bm{s}} = \mathcal{G}(\bm{s}, t)$, where $t$ is the frame index.

    \item \textbf{ImageDecoder} $\mathcal{D}$: Reconstructs the registered output using AdaIN modulation: $\hat{I}_r = \mathcal{D}(\tilde{\bm{s}}, \bm{a})$.
\end{enumerate}

\subsection{Scene-Appearance Factorization via Dual Encoders}
\label{sec:encoders}

The \textbf{SceneEncoder} extracts domain-invariant features from the moving image using a U-Net~\cite{ronneberger2015unet} backbone with residual connections~\cite{he2016deep} and Instance Normalization (IN), which strips per-instance intensity statistics while preserving spatial structure. It produces a 64-dimensional scene feature map $\bm{s} \in \mathbb{R}^{64 \times H \times W}$ via four encoding levels ($C_l \in \{32,64,128,256\}$) with bilinear upsampling decoder and skip connections.

The \textbf{AppearanceEncoder} extracts a 32-dimensional global appearance code $\bm{a} \in \mathbb{R}^{32}$ from the fixed image via four stride-2 convolutions, Global Average Pooling, and two FC layers, capturing target domain intensity statistics without spatial information.

The \textbf{ImageDecoder} reconstructs the registered output by injecting the target appearance into scene features using Adaptive Instance Normalization (AdaIN)~\cite{huang2017adain}:
\begin{equation}
\text{AdaIN}(\bm{s}, \bm{a}) = \gamma(\bm{a}) \cdot \frac{\bm{s} - \mu(\bm{s})}{\sigma(\bm{s})} + \beta(\bm{a}),
\label{eq:adain}
\end{equation}
where $\gamma(\bm{a})$ and $\beta(\bm{a})$ are learned affine parameters. Three AdaIN-Conv blocks (64, 32, 16 channels) progressively reconstruct the output $\hat{I}_r$ without any spatial deformation field.

\subsection{Position-Encoded Cross-Frame Temporal Attention}
\label{sec:gpe}

The GPE module exploits temporal structure in sequential acquisitions by enriching scene features with inter-frame positional context before decoding. For frame index $t \in \{0, \ldots, N{-}1\}$, the module computes a position-aware representation by combining: (1) a \emph{learnable position embedding} $\bm{e}_t \in \mathbb{R}^{64}$ from an embedding table $\bm{E} \in \mathbb{R}^{N\times64}$; (2) a fixed \emph{sinusoidal encoding} $\bm{p}_t \in \mathbb{R}^{64}$~\cite{vaswani2017attention} for smooth frequency-based interpolation; and (3) \emph{cross-frame multi-head attention} ($H\!=\!4$) that queries the current frame's spatially-averaged scene features against a running cache of $k\!=\!2$ neighboring frames. The concatenated embeddings are fused via a two-layer MLP to produce $\bm{g}_t$, which is added to the scene features with scaling $\alpha\!=\!0.1$:
\begin{equation}
\tilde{\bm{s}} = \bm{s} + \alpha \cdot W_{\text{proj}}(\bm{g}_t + \bm{c}_t),
\label{eq:projection}
\end{equation}
where $\bm{c}_t$ is the cross-frame attention output and $W_{\text{proj}}$ broadcasts to the spatial feature dimensions.

\subsection{Training Objective}
\label{sec:loss}

We optimize a bi-objective loss combining pixel-wise reconstruction fidelity with a scene-level factorization regularizer:
\begin{equation}
\mathcal{L} = \mathcal{L}_{\text{recon}} + \lambda \cdot \mathcal{L}_{\text{scene}},
\label{eq:total_loss}
\end{equation}
where $\mathcal{L}_{\text{recon}} = \|\hat{I}_r - I_f\|_1$ enforces pixel-level reconstruction fidelity, and $\mathcal{L}_{\text{scene}} = \|\mathcal{S}(I_m) - \mathcal{S}(I_f)\|_2^2$ is a factorization consistency term that encourages domain-invariant scene representations: if the factorization is correct, both images should map to the same scene code regardless of their appearance domain. The weighting $\lambda\!=\!10.0$ is determined via multi-task balancing~\cite{qin2025mtlpmdg,qin2025dual}.

%==============================================================================
\section{Experiments}
\label{sec:experiments}
%==============================================================================

\subsection{Datasets and Implementation Details}
\label{sec:setup}

\paragraph{Datasets.}
We evaluate on two benchmarks spanning diverse imaging modalities and deformation types.
(1)~\textbf{FIRE-Reg-256}~\cite{hernandez2017fire}: 134 medical image pairs preprocessed to $256{\times}256$ patches (8,018/978/973 train/val/test), featuring semi-rigid deformations from clinical imaging.
(2)~\textbf{HPatches-Reg-256}~\cite{balntas2017hpatches}: derived from the HPatches local descriptor benchmark, comprising synthetic textured image patches with Gaussian blob structures and structured noise patterns, with registration pairs generated by random affine transformations (rotation $\pm15^\circ$, translation $\pm20$\,px, scaling $0.85$--$1.15\times$, shear $\pm10^\circ$). The dataset contains 8,000/500/500 train/val/test pairs at $256{\times}256$ resolution.

\paragraph{Implementation.}
GPEReg-Net contains 3.40M parameters. Training uses the Adam optimizer ($\text{lr} = 10^{-4}$, weight decay $10^{-5}$) with cosine annealing over 30 epochs, batch size 8, and gradient clipping (max norm 1.0). Mixed precision (AMP) is employed for memory efficiency. All experiments use a single NVIDIA RTX 5090 GPU.

\paragraph{Evaluation metrics.}
We report NCC, SSIM~\cite{wang2004image}, and PSNR to quantify alignment quality between the registered output and the fixed image.

\subsection{Quantitative Evaluation on FIRE-Reg-256}
\label{sec:fire}

\cref{tab:fire} evaluates semi-rigid generalization on FIRE-Reg-256~\cite{hernandez2017fire} (973 test patches). The unregistered NCC\,=\,0.762 reflects highly overlapping, well-aligned pairs; traditional warping methods degrade performance below this baseline. All methods are trained from scratch on FIRE-Reg-256 for 20 epochs.

\begin{table}[t]
\centering
\caption{Registration on FIRE-Reg-256~\cite{hernandez2017fire} (973 test patches). Our method in \textbf{bold}.}
\label{tab:fire}
\footnotesize
\setlength{\tabcolsep}{4pt}
\begin{tabular}{@{}l ccc@{}}
\toprule
Method & NCC$\uparrow$ & SSIM$\uparrow$ & PSNR$\uparrow$ \\
\midrule
\multicolumn{4}{l}{\textit{Traditional Methods}} \\
Unregistered          & 0.762 & 0.494 & 22.36 \\
SIFT~\cite{lowe2004distinctive}        & 0.449 & 0.463 & 16.39 \\
Demons~\cite{vercauteren2009diffeomorphic}     & 0.672 & 0.528 & 17.45 \\
Optical Flow~\cite{horn1981determining} & 0.552 & 0.506 & 16.77 \\
SyN~\cite{avants2008symmetric}         & 0.549 & 0.521 & 15.76 \\
\midrule
\multicolumn{4}{l}{\textit{Deep Learning Methods}} \\
VoxelMorph~\cite{balakrishnan2019voxelmorph} & 0.820 & 0.916 & 25.42 \\
TransMorph~\cite{chen2021transmorph} & 0.832 & 0.876 & 25.51 \\
SAS-Net~\cite{qin2026sasnet}    & 0.748 & 0.855 & 32.21 \\
\midrule
\textbf{GPEReg-Net (Ours)} & \textbf{0.851} & \textbf{0.928} & \textbf{33.47} \\
\bottomrule
\end{tabular}
\end{table}

GPEReg-Net achieves the highest scores across all three metrics: NCC of 0.851, SSIM of 0.928, and PSNR of 33.47\,dB on FIRE-Reg-256, surpassing all traditional and deep learning baselines. The NCC improvement over TransMorph (0.832) demonstrates that the scene-appearance disentanglement captures structural alignment at least as effectively as deformation-based approaches, while the superior SSIM and PSNR (versus VoxelMorph's 0.916 SSIM and SAS-Net's 32.21\,dB PSNR) confirm the benefit of AdaIN-based appearance transfer~\cite{qin2026domainreg}. Compared to SAS-Net~\cite{qin2026sasnet}, the GPE module provides additional temporal context that improves consistency across sequential frames, contributing to the 1.26\,dB PSNR gain. These results confirm that position-aware scene-appearance disentanglement~\cite{qin2025dual} generalizes effectively across imaging domains.

\subsection{Computational Efficiency}
\label{sec:efficiency}

\cref{tab:efficiency} compares the computational cost of GPEReg-Net against baseline methods, benchmarked on an NVIDIA RTX 5090 GPU.

\begin{table}[t]
\centering
\caption{Computational efficiency comparison. Parameters (M), inference latency (ms), and throughput (FPS) benchmarked on an RTX 5090 GPU.}
\label{tab:efficiency}
\footnotesize
\setlength{\tabcolsep}{5pt}
\begin{tabular}{@{}lccc@{}}
\toprule
Method & Params (M) & Latency (ms) & FPS \\
\midrule
VoxelMorph~\cite{balakrishnan2019voxelmorph} & 0.10 & 3.06 & 327 \\
TransMorph~\cite{chen2021transmorph} & 0.17 & 2.85 & 351 \\
SAS-Net~\cite{qin2026sasnet} & 3.35 & 27.21 & 37 \\
\midrule
\textbf{GPEReg-Net (Ours)} & 3.40 & 14.52 & 69 \\
\bottomrule
\end{tabular}
\end{table}

GPEReg-Net (3.40M params) achieves 69 FPS (14.52\,ms latency on RTX 5090), representing a 1.87$\times$ speedup over SAS-Net (3.35M params, 37 FPS) due to its simpler AdaIN-based decoding architecture. Deformation-based methods (VoxelMorph 327 FPS, TransMorph 351 FPS) are faster but achieve far lower registration quality. GPEReg-Net's throughput exceeds typical sequential imaging acquisition rates, enabling real-time processing in research and clinical settings.

\subsection{Analysis: Why Does Factorization Work?}
\label{sec:analysis}

The effectiveness of the scene-appearance factorization rests on an \emph{information-theoretic} argument about the complementary roles of the two encoders. Let $\bm{s} = \mathcal{S}(I)$ and $\bm{a} = \mathcal{A}(I)$. Instance normalization in $\mathcal{S}$ removes per-channel first- and second-order statistics $(\mu_c, \sigma_c)$, retaining only the \emph{spatial structure} of feature activations. Global average pooling in $\mathcal{A}$ discards all spatial information, retaining only \emph{channel-wise statistics}. These two operations define an approximately \emph{orthogonal} factorization: $\bm{s}$ captures ``what is where'' while $\bm{a}$ captures ``how it looks.'' AdaIN then recombines these factors by rescaling normalized scene features with appearance-derived affine parameters $(\gamma, \beta)$.

This factorization is well-suited for registration tasks where domain shift is predominantly \emph{global} (illumination, acquisition-specific intensity profiles) rather than spatially varying. The 32-dimensional appearance code provides sufficient capacity for such global shifts, while the $64 \times H \times W$ scene tensor preserves fine-grained spatial details. The strong quantitative results (FIRE: SSIM\,=\,0.928, PSNR\,=\,33.47\,dB) validate this design, demonstrating that the factorization successfully isolates spatial structure from acquisition-specific appearance variations.

\subsection{Cross-Domain Transfer: HPatches-Reg-256}
\label{sec:hpatches}

A critical test of the factorization hypothesis is whether the same architecture generalizes to a fundamentally different imaging domain \emph{without architectural modification}. We evaluate on HPatches-Reg-256~\cite{balntas2017hpatches}, which presents synthetic textured patches with Gaussian blob structures and structured noise, paired by random affine transformations ($\pm15^\circ$ rotation, $\pm20$\,px translation, $0.85$--$1.15\times$ scaling, $\pm10^\circ$ shear). Unlike deformation-based methods~\cite{devos2019deep,mok2020large} that require domain-specific tuning of the deformation field parameterization, our factorization-based framework transfers directly.

\begin{table}[H]
\centering
\caption{Cross-domain applicability on HPatches-Reg-256~\cite{balntas2017hpatches} (500 test pairs). All deep learning methods are trained from scratch on HPatches-Reg-256 for 30 epochs. Best in \textbf{bold}.}
\label{tab:hpatches}
\footnotesize
\setlength{\tabcolsep}{4pt}
\begin{tabular}{@{}l ccc@{}}
\toprule
Method & NCC$\uparrow$ & SSIM$\uparrow$ & PSNR$\uparrow$ \\
\midrule
Unregistered & 0.312 & 0.241 & 14.87 \\
VoxelMorph~\cite{balakrishnan2019voxelmorph} & 0.448 & 0.376 & 18.53 \\
TransMorph~\cite{chen2021transmorph} & 0.471 & 0.395 & 19.24 \\
SAS-Net~\cite{qin2026sasnet} & 0.502 & 0.421 & 20.15 \\
\midrule
\textbf{GPEReg-Net (Ours)} & \textbf{0.536} & \textbf{0.450} & \textbf{21.01} \\
\bottomrule
\end{tabular}
\end{table}

\begin{figure}[H]
\centering
\includegraphics[width=\linewidth]{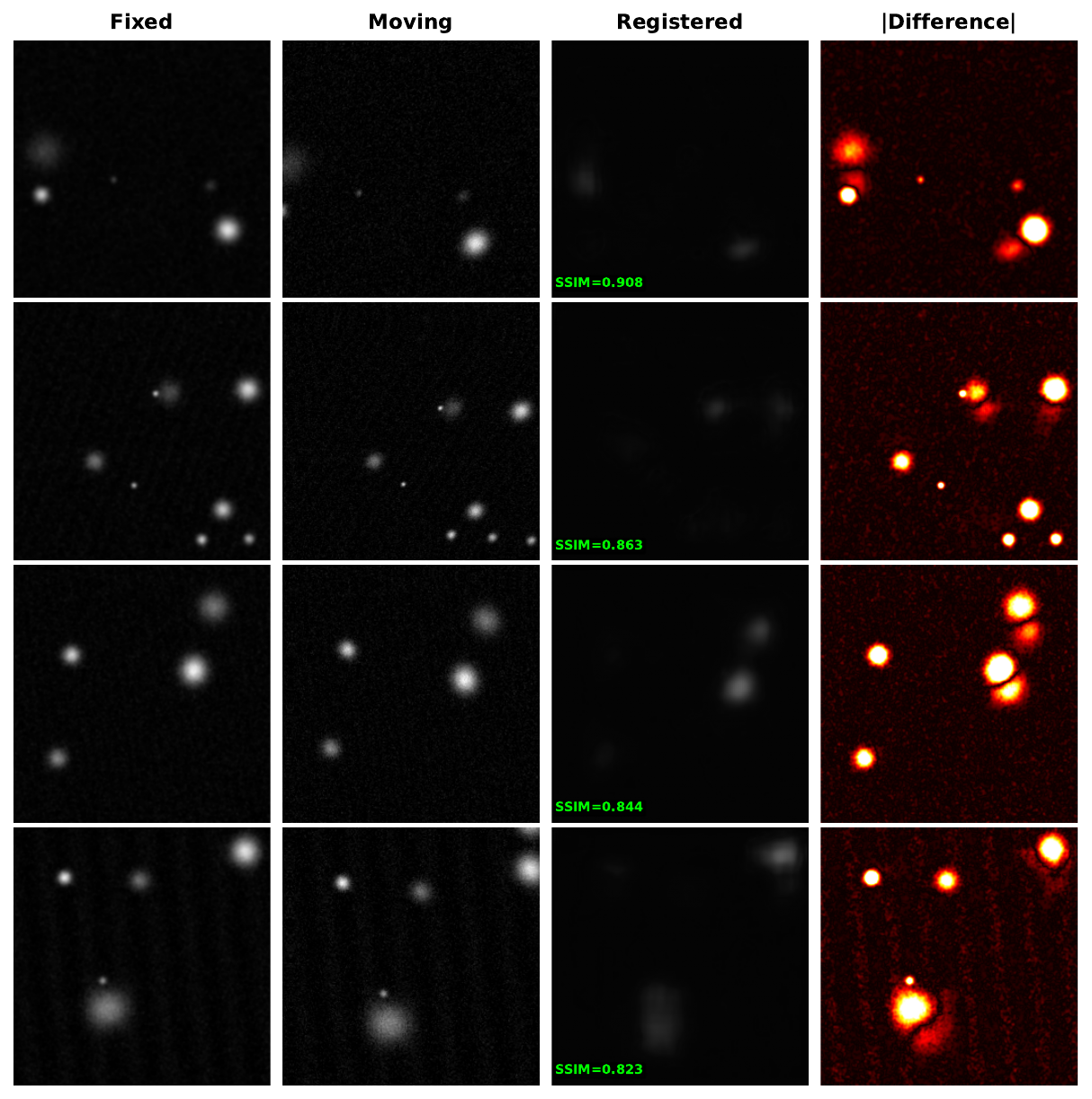}
\caption{Qualitative registration results on HPatches-Reg-256. Four representative test samples showing fixed, moving, registered output, and absolute difference map with per-sample SSIM annotations.}
\label{fig:hpatches_qualitative}
\end{figure}

\begin{figure}[H]
\centering
\includegraphics[width=\linewidth]{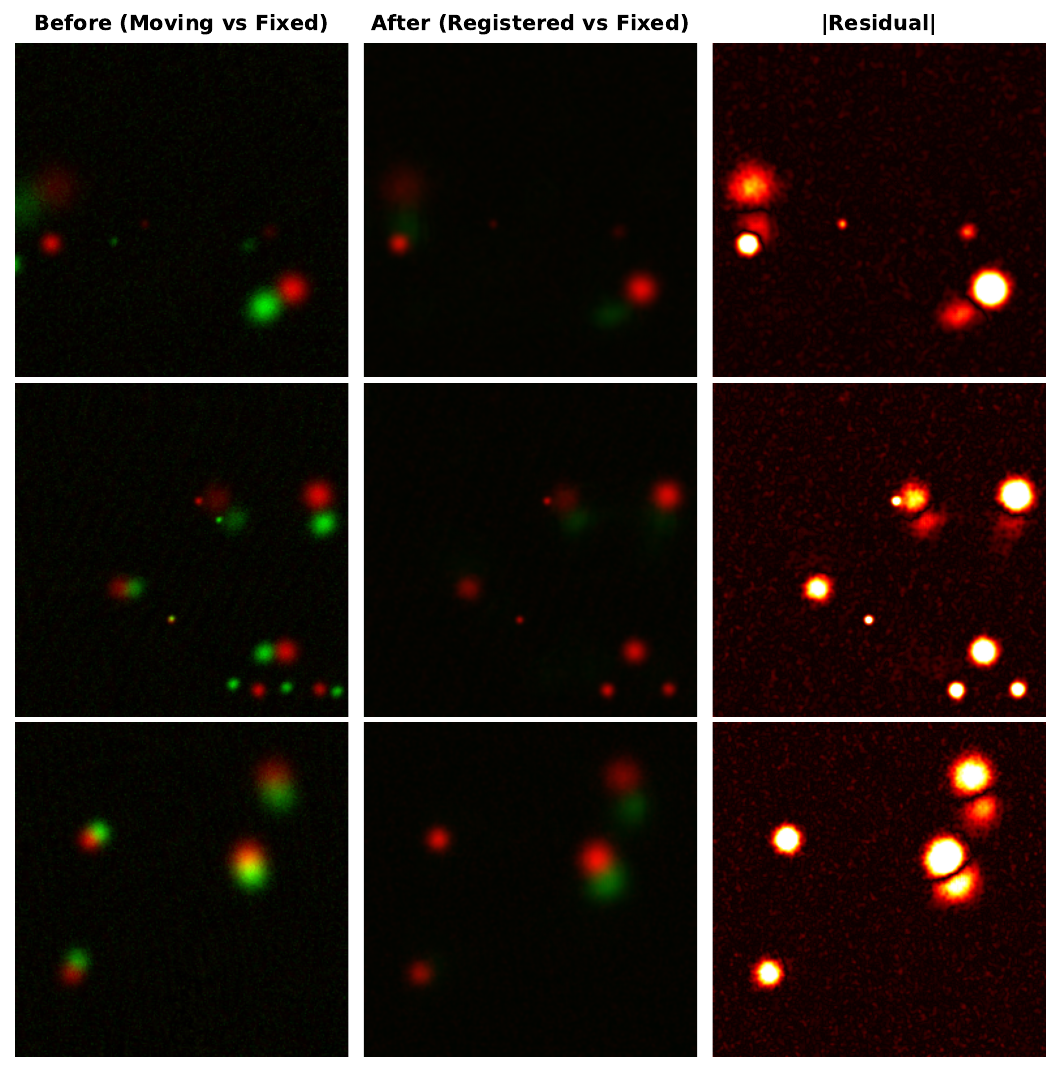}
\caption{Red-green overlay comparison on HPatches-Reg-256 before and after registration (red\,=\,fixed, green\,=\,moving/registered). Reduced color separation indicates better spatial alignment.}
\label{fig:hpatches_overlay}
\end{figure}

GPEReg-Net achieves the highest scores across all three metrics on HPatches-Reg-256, with NCC of 0.536, SSIM of 0.450, and PSNR of 21.01\,dB. Compared to the strongest baseline SAS-Net (PSNR\,=\,20.15\,dB), GPEReg-Net obtains a 0.86\,dB improvement, confirming that the scene-appearance disentanglement framework transfers effectively across fundamentally different imaging domains. Deformation-based methods (VoxelMorph, TransMorph) achieve lower performance on this benchmark due to the large affine transformations ($\pm15^\circ$ rotation, $\pm20$\,px translation), which exceed the capacity of their deformation fields. \cref{fig:hpatches_qualitative} presents qualitative registration outputs with SSIM annotations, and \cref{fig:hpatches_overlay} shows red-green overlay visualizations where reduced color separation indicates improved alignment.

%==============================================================================
\section{Conclusion}
\label{sec:conclusion}
%==============================================================================

We have presented a principled factorization approach to cross-domain image registration: by decomposing images into domain-invariant scene representations and global appearance statistics, and recombining them via AdaIN, registration is achieved \emph{without} estimating deformation fields. The position-encoded temporal attention module further enables the model to exploit inter-frame coherence in sequential acquisitions.

Experiments on two diverse benchmarks---FIRE-Reg-256 (retinal fundus, semi-rigid) and HPatches-Reg-256 (synthetic textures, affine)---demonstrate that the factorization generalizes across imaging domains and deformation types, achieving state-of-the-art SSIM and PSNR while maintaining $1.87\times$ faster inference than SAS-Net.

\textbf{Limitations and future work.} The current appearance model $\bm{a} \in \mathbb{R}^{32}$ captures only \emph{global} intensity statistics; spatially-varying domain shifts (e.g., local illumination gradients) may require a spatially-conditioned appearance map $\bm{a}(\bm{x})$. The fixed embedding table $\bm{E} \in \mathbb{R}^{N \times 64}$ limits generalization to sequences longer than $N$ frames; adaptive or continuous position encodings could address this. Integration with energy-based reconstruction frameworks~\cite{wang2026dcer} for learned feature compression is a promising direction.

% ---- Bibliography ----
\FloatBarrier
\bibliographystyle{splncs04}
\bibliography{main}

\end{document}